# EssentialGIN: a new approach for gene essentiality prediction based on graph isomorphism neural networks

Sahar Mansourirad[1], Zahra Narimani[1]*, Parvin Razzaghi[1], Nazanin Hosseinkhan[2]

[1] Department of Computer Science and Information Technology, Institute for Advanced Studies in Basic Sciences (IASBS), Zanjan 45137-66731, Iran

[2] Endocrine Research Center, Institute of Endocrinology and Metabolism, Iran University of Medical Sciences, Tehran, Iran

* corresponding author: narimani@iasbs.ac.ir

## Abstract

**Background:** Prediction of essential genes (proteins), is a basic and challenging problem but at the same time very costly and time-consuming in wet-lab experiments. Predicting essential genes, only based on computational methods (to introduce wet-lab candidates) using centrality measures are not accurate and result in large number of false positives; therefore, more complex models such as deep learning and also integration of biological information are used in recent research to identify essential genes.

**Methods:** In this work we focus on graph isomorphism networks, in order to embed proteins as a node in PPI network to conserve topological features of PPI network, and also integrate biological data such as gene expression data, gene orthology information and gene subcellular localization information, and introduced a deep architecture for predicting essential genes. Graph isomorphism network architecture is modified in this work for embedding node information.

**Results:** Our experiments proved that the proposed method outperforms baseline centrality-based methods and also machine learning based methods such as Node2Vec, MLP, and also graph attention networks (GAT).

**conclusion:** In this paper we observed that using graph isomorphism networks that integrate biological data (as node attributes) and preserve network topology can significantly improve the essential gene prediction accuracy. In simpler organisms such as E. coli and D. melanogaster, methods such as multi-layer perceptron using Node2Vec embedding also performs very good, but in H. sapiens the introduced architecture significantly outperforms deep learning and other graph neural network solutions.

## Background

Data with an underlying graph structure is prevalent across various scientific domains, including biology, social sciences, and telecommunications. Social networks, as one of the most prominent examples of such structured data, have been the subject of extensive research. Tasks such as community detection, detecting nodes with high centrality, and link prediction are typical when dealing with these datasets. Similarly, in bioinformatics, data such as protein-protein interaction networks, gene regulatory networks, drug-target interaction networks, and disease-drug correlation networks are available in different problems. Prediction of essential genes (proteins), which are indispensable for cellular survival, is a basic and challenging problem in understanding fundamental biological processes in a cell. These genes are crucial targets in therapeutic strategies, such as in chemotherapy for cancer treatment, where they serve as focal points for drug action to eradicate cancerous cells, or in assessing the side effects of pharmaceuticals. Proteins seldom operate in isolation within a cell; they perform their cellular functions in collaboration with a complex network of other proteins. Laboratory methods, such as single-gene knockdown, RNA interference, and transposon mutagenesis, are available for identifying essential genes and proteins in living cells. However, these methods are often prohibitively expensive and time-consuming. For instance, single-gene knockout (deletion) experiments involve removing a gene—the fundamental unit of protein synthesis—and observing whether the cell can survive without it. If the cell does not survive, the deleted gene or its protein product is considered essential. However, the vast number of genes present (thousands, even in the simplest organisms) makes the exhaustive laboratory examination of each gene impractical. Therefore, employing computational methods to predict essential proteins and generate a list of candidate genes for subsequent laboratory validation represents a significant advancement in reducing both the costs and duration of experimental research [1].

Considering the graph structure underlying protein-protein interactions, one of the primitive computational approaches toward identifying essential genes is through using centrality measures. Centrality measures are quantitative indices that evaluate a vertex's significance within a network based on various structure-related criteria, such as the number of adjacent vertices, the density of edges in their neighboring, and the betweenness of a vertex in connecting other nodes/graph components. Using centrality measures, we can nominate those proteins with elevated centrality as prime candidates for essentiality [2-4]. Nonetheless, essential protein prediction methods based only on centrality metrics depend exclusively on the network's topological characteristics and are insufficient for the accurate prediction of essential genes. A drawback of using only network-based metrics to predict essential genes, is the high number of false positive or false negative interactions within existing protein interaction networks, which are often resulting from using techniques that may not accurately represent genuine protein relationships since these interactions are approved in lab experiments – and not validated in a living cell, and might omit certain protein interactions, or report non-existing ones. Additionally, the essentiality of a gene is often related to a variety of biological factors rather than just the network's topological traits. More comprehensive insights into centrality metrics for forecasting essential genes/proteins are available in a review paper by Li et al. [5].

An alternative strategy employs machine learning techniques to integrate biological features extracted from various biological data, aiming to provide robust evidence for predicting the essentiality of proteins. Zhang and colleagues conducted a survey of machine learning-based predictive methodologies for essential genes, emphasizing network topological attributes and highlighting prospective challenges and avenues for future research. [6]. In the conventional machine learning paradigms, the selection and extraction of features—typically inclusive of centrality measures—are manually conducted, which requires pre-existing domain expertise and an intricate comprehension of the interplay between gene essentiality and the spectrum of biological data, to harvest attributes that are beneficial for model training, albeit the accuracy of such pre-existing knowledge is not guaranteed [7]. Recently, other methods are hired to extract node features in biological networks, for example meta-path based network analysis [8] and hypergraph models [9] are used in disease-gene association studies. Random forest based methods has also been applied for detecting essential genes in E. coli by Wang et. al. [10].

Recent advancements in deep learning have significantly enhanced bioinformatics applications, such as medical image segmentation and pharmacological target prediction [11], and the identification of vital genes [12-14]. Convolutional neural networks (CNNs) have revolutionized the automated extraction of features from medical images [15] and genomic sequences [11]. Zheng and colleagues harnessed CNNs to extract functional patterns from gene expression time-series data, subsequently transforming these patterns into visual representations aligned with cellular cycles [12]. Moreover, they leveraged bidirectional long short-term memory (LSTM) networks for feature extraction from gene expression profiles within the same dataset, aiming to ascertain gene essentiality. Zeng et al. [13] integrated gene expression data, intracellular localization data, and protein-protein interaction networks (a Node2Vec embedding) are integrated to predict essential genes in Saccharomyces cerevisiae. Hassan and colleagues implemented a six-layer neural network for essential gene prediction in microorganisms, predicated only based on genomic sequence data, with manual feature extraction [14].

In the current study, we propose a graph neural network-based approach for identifying essential genes, which in addition to using biological knowledge such as gene subcellular localization and orthology data, relies on a specific type of graph neural networks, called graph isomorphism network (GIN), for integrating PPI information. In the rest of this paper, in previous work methods, we provide more detail on graph representation-based methods for solving the current problem, and mention the application of graph neural network-based architectures – also their pros and cons – used in existing research. Then we will discuss why graph isomorphism networks are preferable architecture for modeling PPI networks in finding essential genes. Then the proposed method is described and result and discussion part are presented.

## Related work

In this part we focus on existing essential gene detection methods based on graph representation learning. A novel deep learning-based graph embedding technique was proposed for autonomously learning low-dimensional representations of vertices [16]. The exploration of random walks and metapath extractions in the lncRNA-Protein interactive network was undertaken [17]. Zheng et al. have contributed a notable study employing the node2vec algorithm [16], which innovated a deep learning-driven network embedding technique to extract features for proteins from protein-protein interaction networks. Their findings revealed that such low-dimensional feature representations surpass manually derived centrality measures in informational richness [12], [13]. Wang et al. [18]

implemented a robust weighted network approach to enhance the prediction of gene essentiality, by adding weights for representing inaccuracies such as false positives or negatives within protein-protein interact. Furthermore, they integrated this weighted network with supplementary biological datasets, encompassing gene expression profiles, subcellular localization, and orthology data, to refine their predictive model.

Zhang et al. [19] introduced DeepHE, an innovative deep learning framework designed for predicting essential genes. In DeepHE, features derived from DNA and protein sequences are integrated with an embedding vector deduced from protein-protein interaction networks through the application of the node2vec algorithm. The framework employs a deep neural network, which, is designed to tackle with the imbalances inherent in class distributions, boasts predictive capabilities with an AUC-ROC exceeding 94%, AUC-PR surpassing 90%, and an overall accuracy above 90%. It ingests dual-modal data inputs, encompassing both sequence and protein-protein interaction network data. During the feature extraction phase, a multitude of sequence attributes are extracted from each gene's nucleotide and protein sequences.

The node2vec approach is leveraged to extract semantic features relevant to each gene from the network. Finally, these feature vectors are concatenated to serve as inputs for a classification module. This module is implemented using a multi-layer neural network architecture, comprising with several fully connected hidden layers and a terminal output layer. The hidden layers uniformly employ the ReLU activation function, while the output layer utilizes the sigmoid function to execute classification tasks. The network's cost function is articulated as binary cross-entropy. Posterior to each hidden layer, a dropout layer is strategically deployed to diminish the network's susceptibility to training data noise, thereby augmenting its generalization capacity.

To address DeepHE shortcomings such as obfuscating valuable biological insights [20], the authors augmented DeepHE with additional features encompassing gene ontology, protein complex characteristics, and domain-derived attributes, and proposed DeepSF network [19]. In DeepSF, engineered features such as the inverse domain frequency (IDF) sum and the S score, defined as $S = |M|/|N|$, indicative of the proportion of a gene's neighbors engaged within a protein complex. Subsequently, the ensemble of features, inclusive of those generated by node2vec, is fed into the deep learning model to ascertain gene essentiality. This approach boasts the capability to predict human gene essentiality with an average AUC-ROC of 95.17%, AUC-PR of 92.21%, and an F1 score around 78.71%.

Despite the efficacy of these deep learning strategies, they necessitate the employment of a vertex embedding technique, such as node2vec, to convert protein-protein interaction networks into a standardized and consistent input format. Since the topology of a network should be embedded in a numerical vector to be used in a neural network input, these methodologies use a representation extracted from the input PPI networks, thereby neglecting a substantial part of the graph's topological data, which can lead to suboptimal performance in scenarios involving intricate graph structures.

Schapke et al. proposed EPGAT, an innovative computational technique predicated on Graph Attention Networks (GAT) for the prediction of gene essentiality [21]. Experimental results show that EPGAT performs superior to network-based method (DC, LAC, NC) significantly, and also performs better or comparable (in H. sapiens) to other machine learning methods (SVM, MLP, N2V). Graph Attention Networks harness the principles of generalized convolutional neural networks [22] and are tailored for extracting important graph structural features. This technique enables extracting local features regarding graph vertices via self-attention mechanisms [23], thus facilitating the vertices to Perceive the relative significance of each neighbor throughout the aggregation phase. In EPGAT, gene expression profiles, subcellular localization data [24], and orthology information are integrated as features of the graph vertices, and in addition, Graph Attention Networks are also used in order to consider important topological features; therefore biological information and graph topological features are very well leveraged in order to predict protein essentiality [20]. In the current work, we use the same intuition of integrating biological knowledge and benefit from topological features. As a result, we propose a deep architecture for vertex classification, which is based on graph isomorphism networks (GIN). GINs preserve graph structure information by leveraging graph isomorphism principles. Our intuition was based on the fact that existing graph neural network methods proposed for this problem are not based on GIN, and also existing research show the role of structural information in predicting gene essentiality. We adopted the architecture of GIN in order to use it for vertex classification (i.e. essential gene prediction), also used gene orthology information, gene expression data, and gene subcellular localization information. Having all of this information, our experimental result demonstrated improved performance compared to Node2Vec, MLP, EPGAT, and the baseline degree centrality measure for predicting essential genes.

## Method

The performance of deep learning-based methods on graph data has demonstrated their superiority in research [25]. On data with graph structure, Graph Neural Networks (GNNs) have shown superior performance and been known as state-of-the-art method in many research problems. However, GNNs have a known limitation in considering graph topological features in the embedding they generate. Graph Isomorphism Network, is a network architecture for embedding graph structured data, that has shown its strength to capture graph structure [26]. Knowing that the structural properties of PPI networks have been observed to give evidence about gene essentiality, our intuition is that using GINs can be beneficial in gene essentiality prediction.

The Graph Isomorphism Network (GIN), is a type of deep learning network that uses several techniques to increase the classification of graphs in graph neural networks. These networks obtain node embeddings in several stages and provide the output of each hidden layer to the next hidden layer to create a new vector. GIN needs the iterations to be applied on the whole graph simultaneously, since each node uses the embeddings of all nodes from previous layer. The representation of a node v in layer k of GIN is computed using (1):

$$h_v^{(k)} = MLP\left(\left(1 + \varepsilon(k)\right). h_v^{(k-1)} + \sum_{u \in \mathcal{N}(v)} h_u^{(k-1)}\right) \quad (1)$$

In which, $h_v^{(k)}$ is the representation of node v at layer k, $\mathcal{N}(v)$ are neighbors of node v, ε(k) determines the importance of the target vertex compared to its neighbors (if their importance is the same, ε=0). ε can be a learnable parameter (ε(k)) or a constant scalar. The graph embedding vector layer is obtained by concatenating the node embeddings produced in the GIN. To obtain the graph embedding, the vertex descriptor vectors can also be aggregated using mean, sum, max functions (2-4) or through concatenating node embeddings (5).

$$Mean{:}\ h_G = \frac{1}{N}\sum_{i=0}^{N} h_i \quad (۲)$$

$$Sum{:}\ h_G = \sum_{i=۰}^{N} h_i \quad (۳)$$

$$Max{:}\ h_G = max_{i=0}^{N}\ (h_i) \quad (۴)$$

$$h_G = concatenate(h_0, h_1 \dots, h_n) \quad (5)$$

The graph embedding, $h_G$, is used in a classifier to classify the graphs. This method has been suggested for graph classification, but in our current research, we make some modifications to it to convert it into a method for node classification and we will use it to distinguish essential proteins from non-essential ones.

Considering that it has been proven that the Graph Isomorphism Network is theoretically equivalent to the Weisfeiler-Lehman test for distinguishing graphs [27], and also considering the reviews conducted in the article [30], it can be concluded that the Graph Isomorphism Network provides a powerful model for classifying graphs. We also used this idea and converted this model into a model for node classification. For this purpose, inspired by these types of networks, we made the following changes in the Graph Isomorphism Network (GIN) so that it can be used for vertex classification.

For designing the vertex descriptor layer, which we call GINConv here, based on the idea proposed in [28], we consider each layer to include a linear layer, a normalization layer, a ReLU activation function, another linear layer, and again a ReLU activation function. Fully connected layers, also known as linear layers, connect each input neuron to each output neuron. Activation functions are for determining whether the vertices of the graph, which are in fact our neural network neurons, should be activated or not. But the problem is that after each hidden layer, the distribution of input data changes, which slows down learning, which is called internal covariate shift. To solve this problem, we use a batch normalization layer (Figure 1). With each normalization, the mean becomes zero and the variance of the data becomes one.

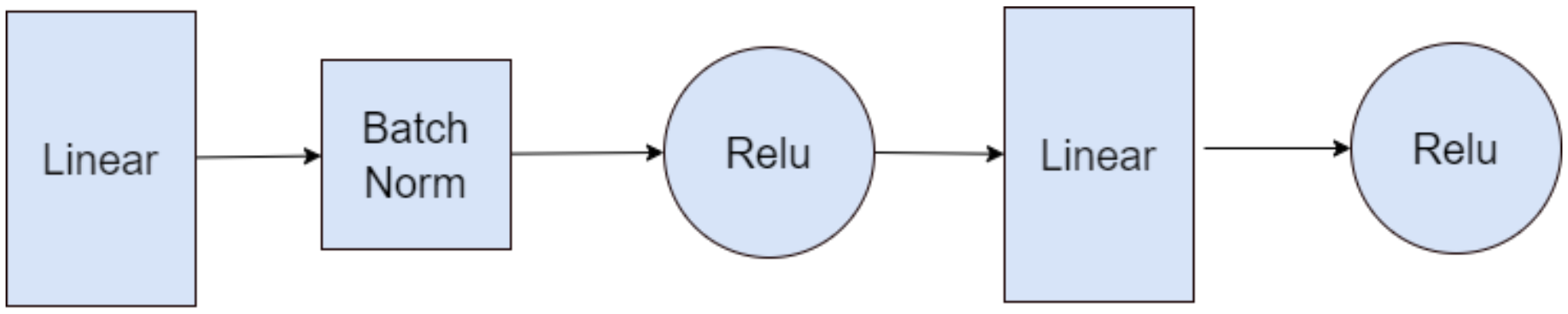

Figure 1- Design of the vertex descriptor layer or GINConv

In the article [26], the number of these layers is five, but in the proposed method, we use four layers, as the experimental results show that the performance of five and four layers is almost identical. In GIN, the last layer is used to obtain the graph embmedding using the output of the GINConv layers, while we only need the vertex embedding vectors, therefore, by removing the GIN last layer, we will have vertex embedddings. After this, we give the output of the concatenation of the previous layers to the classifier. In this research, in the experiments, we added a dropout layer after each GINConv layer and got a better result. Eventually, we will have a model as shown in Figure 2. The final model of the proposed method is shown in Figure 2, and we name it GIN*.

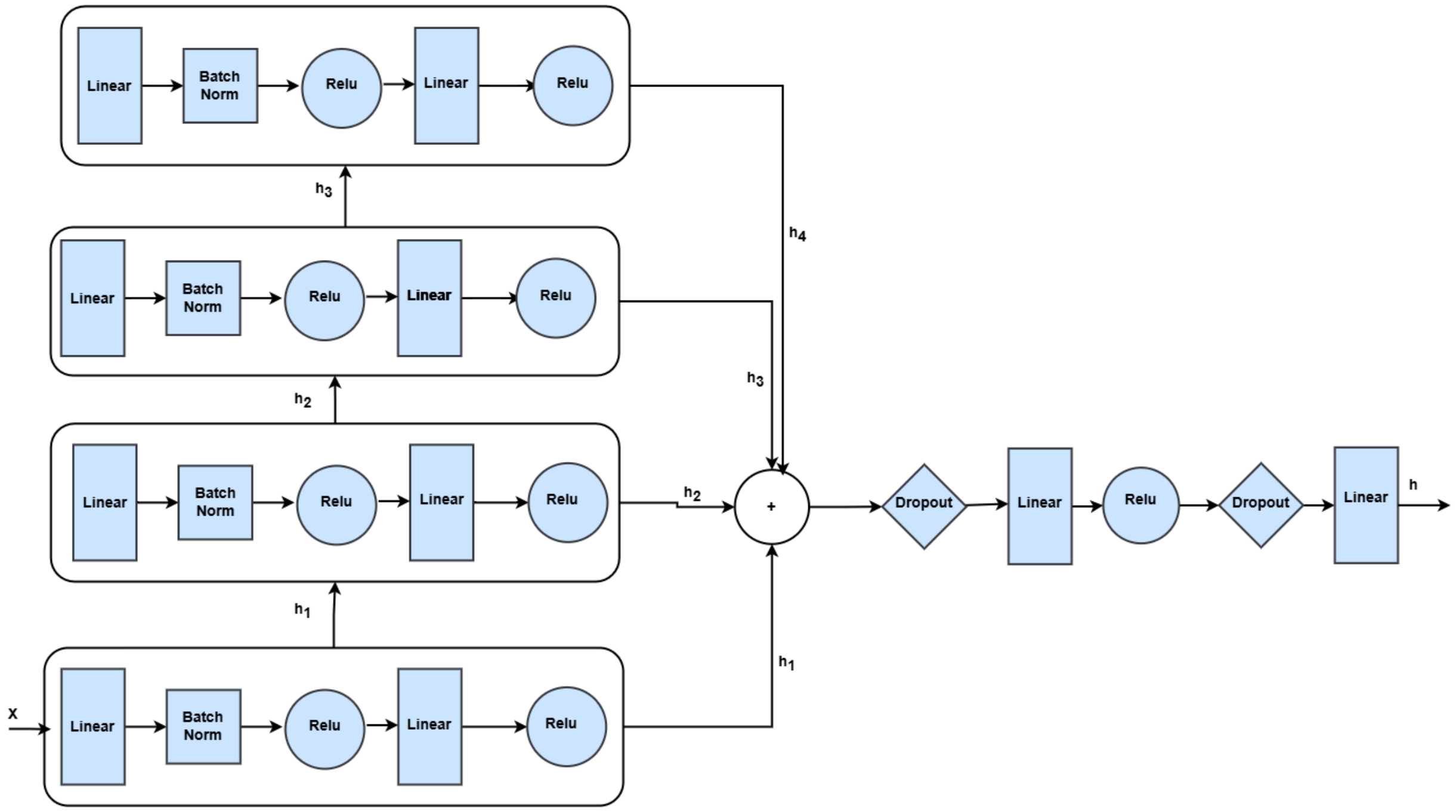

Figure 2. Proposed modified GIN (GIN*) structure for learning node embeddings

In the final proposed architecture, we use a combination of the modified GIN, GIN*, and MLP. In such a way that we first present the existing data to the GIN*, and provide the output of this part alongside other biological data to an MLP network. The final deep network, which we have abbreviated as EssentialGIN, is as shown in Figure 3. The best dropout rate reported in [20] has been obtained for different species. The parameter settings for the learning rate, weight decay, dropout rate for GINConv, and dropout rate for the classifier in the proposed model are respectively 0.05, 5e-4, 0.1, and 0.3.

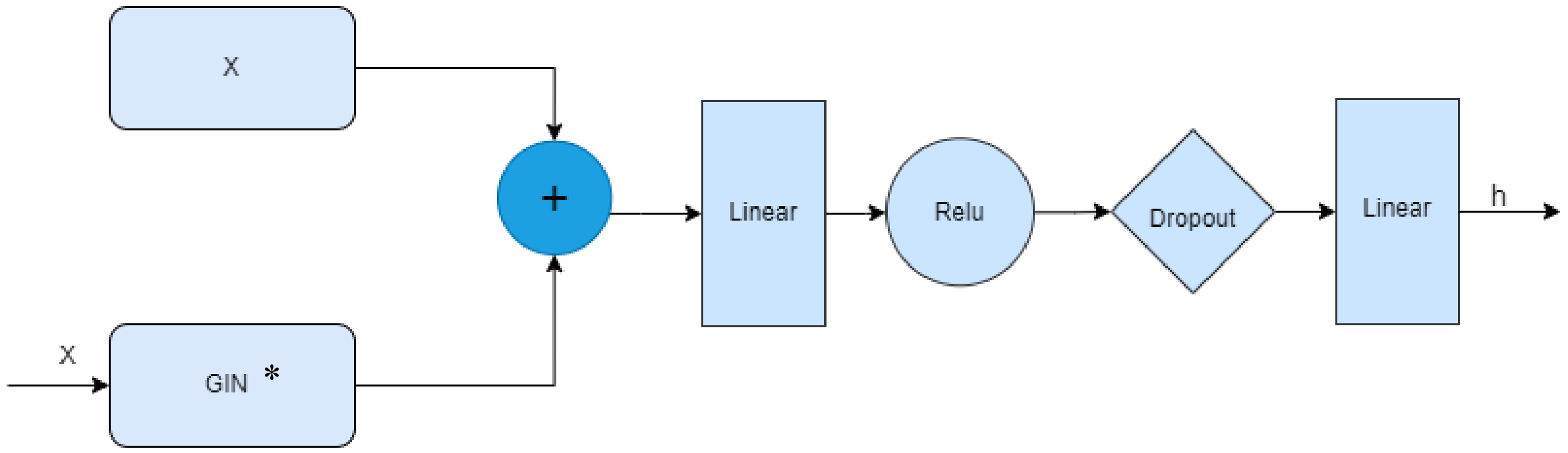

Figure 3- Structure of the ESSENTIALGIN deep neural network in the proposed method (EssentialGIN)

For training and evaluating the model, the labeled data was randomly divided into a training set (80%) and a test set (20%) using stratified division. To better estimate the power of the models, ten random divisions were used, and the average of the performance metrics obtained during the iterations was taken. In addition, we used a subset of the training dataset for validating the proposed model during training and for early stopping of the training process. All codes are in Python and available on our github[1]. For learning and evaluation, as well as for MLP models, PyTorch was used, and for graph-based models, the Pytorch Geometric extension was used. Since in each random execution of the method, the best results occurred at epoch 500, in all experiments, the number of epochs for the GIN part of the network was set to 500, and for the second part of the network, i.e., MLP, considering the cost function values and accuracy, the number of epochs was set to 1000. Also, in all experiments, with the aim of reducing false positive interactions, connections with a confidence score of less than 0.5 were filtered for the STRING protein-protein network. It should be noted that all codes were run on Google Colab and the results are presented in Tables 5-7. As the results show, all methods, when we add gene expression data to the PPI network, showed their best performance.

## Results and discussion

As baseline methods, this study adopts approaches from four categories found in the related work: Multilayer Perceptron (MLP) which uses non-deep graph embedding Node2Vec as input, random forest, support vector machine (SVM) and gradient boosting from the category of machine learning algorithms (traditional machine learning), Degree Centrality (DC), clustering coefficient and betweenness centrality from structure-based methods, and deep learning-based method EPGAT. In these algorithms, standard feature vectors are created for each gene using information such as gene expression data, subcellular location information, and gene orthology data and integrating this information with the topology of the protein-protein interaction network. Since integrating biological data such as gene subcellular localization and gene expression data into the input had a significant effect on the result, we represented the final results in different rows regarding which data is used. Comparisons with N2Vec, MLP, degree centrality and EPGAT are provided in result section (with different input data integration), and comparisons with betweenneess centrality, clustering coeffeicient, SVM, random forest and gradient boosting for H. sapiens are provided in tables A1-A3 in appendix A.

In this study, an MLP network with a hidden layer of 32 neurons was used and trained with a dropout rate of 0.2. In our MLP model, network features are extracted using the node2Vec (N2V) embedding method – referred as N2V-MLP in the results. N2V uses random walks on graphs to learn node embeddings. The parameter values for N2V were empirically obtained based on the best settings from analyzing multiple experiments with different configurations. We used these vectors as additional features for the MLP training set to ensure a fair comparison. The dimension size of the node embeddings was set to 128 and the random walk length, the number of walks per node, and the context size were set to 64 [18]. All other parameters were applied with their default values.

According to the experiments conducted in this study (Table 3-5 in the result section), the N2V-MLP algorithm provides significantly better results compared to DC, MLP and EPGAT on model organisms. However, this algorithm uses the powerful but relatively time-consuming N2V method. The N2V-MLP network architecture is represented in Figure 4 (using the implementation available in [20]).

[1] https://github.com/saharmansourirad/EssentialGIN/

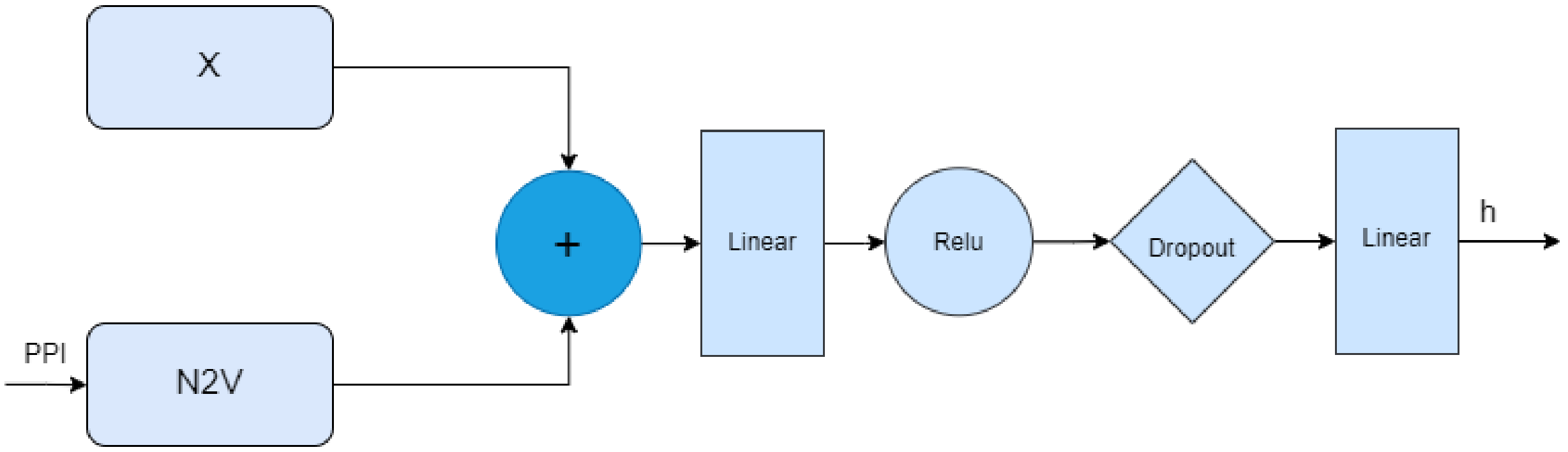

Figure 4. the architecture of N2V-MLP model used in the experiments

One of the other methods used for comparison purposes is EPGAT [20] which is based on Graph Attention Networks (GAT) as it is described in the previous work section. Parameter setting in EPGAT in our experiments is as follows; learning rate = 0.5, weight decay = 5e-4, number of hidden units = 8, number of attention nodes = 8, and random dropout = 0.4.

## Datasets

We used four types of data for evaluating the proposed method; PPI networks, gene expression profiles, subcellular localization, and gene orthology information. Figure 5, displays the final model and the input data. We used UniProtKB ids for encoding genes and proteins, and those entities with no id in UniProtKB were excluded. Essential (and non-essential) genes were extracted from Online GEne Essentiality (OGEE) databse [29]. 21,556 genes (8.9% essential) for H. Sapiens were identified, which was reduced to 18,476 (9.88% essential) genes after standardization and UniprotKB encoding. Our experiments are evaluated using three PPI datasets; DIP [30], BioGRID [31], and STRING [32]. Table 1 provides a description of these datasets for human genes.

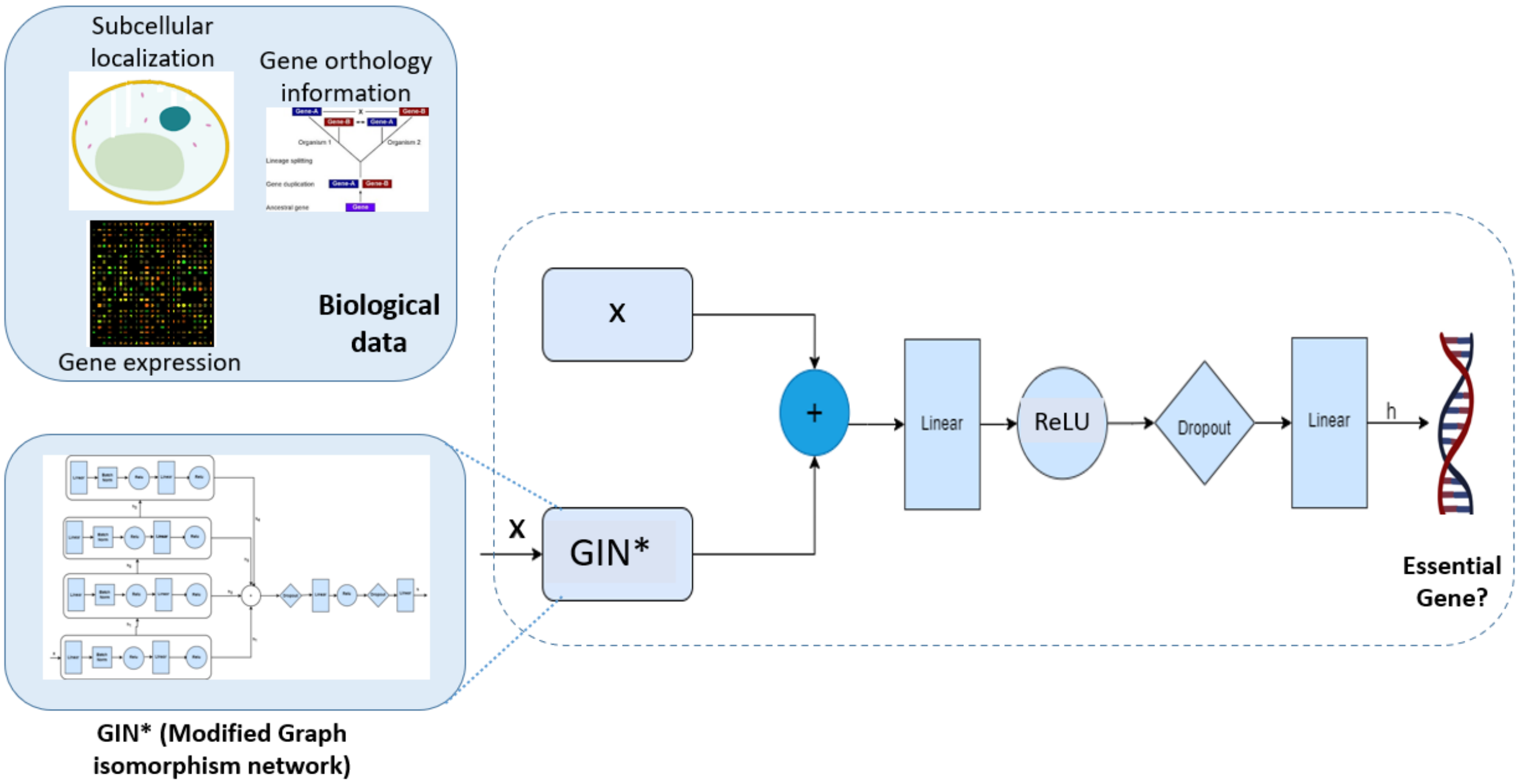

Figure 5. the architecture of the proposed methods (EssentialGIN); x represents the input data which can optionally be gene expression, subcellular localization and gene orthology information. GIN* (see fig 2 for the detail of GIN*) is the modified GIN for node classification. The input graph structure based on which GIN* embeds node information is the selected PPI (STRING, DIP, or BioGRID)

**Table 1. attributes of different PPI datasets used for evaluation purposes (for H. Sapiens)**

| Species | H. Sapiens | | |
|---|---|---|---|
| PPI network | BioGRID | STRING | DIP |
| #nodes | 16562 | 18822 | 4615 |
| #edges | 479496 | 1340788 | 7417 |
| #labeled_genes | 14486 | 17894 | 3370 |
| #essential_genes | 1590 | 1806 | 726 |
| #test_genes | 2898 | 3579 | 674 |

Since the number and also types of interactions are different in these PPI networks, the results of gene essentiality prediction will be dependent on the selected network. The performance of the proposed EssentialGIN method is reported separately for each network.

As mentioned, three other attributes are used in the input; namely subcellular localization, gene expression profiles and gene orthology information. Essential genes are known to be located in inner cellular components more than cell membrane [33], and subcellular information can therefore be useful in increasing the performance of gene essentiality prediction. We extracted subcellular information from COMPARTMENTS database [34]. For retrieving orthology information [35], the InParanoid [36] dataset is used. Each gene will be represented with a vector of fixed length in the input. The components of this vector are described in table 2.

**Table 2. the number of features related to gene expression, orthology and subcelluar localization information in the representing vector for each gene**

| attributes | Gene expression data | Orthology information | Subcellular localization |
|---|---|---|---|
| Number of features for H. sapiens representing gene vector | 64 | 162 | 11 |

The gene expression profiles are measurements representing gene activity in different timepoints or specimens [20]. We used gene expression data 67547GSE [37] extracted from GEO [38] – in order to be compatible to our baseline methods.

We split the dataset to train (80%) and test (20%) partitions using stratified sampling in order to preserve the label in order to maintain the proportions of positive/negative labels. We used 10-fold cross-validation and the average of 10 run is reported. A subset of training set is used as validation set in order to apply early-stopping. The code is implemented in Python – Pytorch and Pytorch Geometric (for models based on Graph) libraries. For GIN, 500 epochs and for MLP 1000 epochs are used (determined experimentally). The interactions with confidence less than 0.5 from STRING dataset are excluded from the input (similar to baseline methods).

The evaluation criteria in the following experiments include AUROC (area under ROC curve) and AUPR (area under precision-recall curve) available measures in EPGAT paper to compare with this algorithm, and precision, recall, Matthews Correlation Coefficient (MCC) (eq. 9), and accuracy (eq. 10) in other experiments (not including EPGAT).

$$\mathrm{MCC} = \frac{TP \times TN - FP \times FN}{\sqrt{(TP + FP)(TP + FN)(TN + FP)(TN + FN)}} \quad (9)$$

$$Accuracy = \frac{TP + TN}{TP + FP + TN + FN} \quad (10)$$

Accuracy relies on the total number of correctly classified samples; and that's why accuracy is relatively high in all the experiments. MCC is a more realistic measure in imbalance scenarios. We performed all of our experiments (in the main text and appendix) on different PPI networks DIP, STRING and BioGRID. In Tables 3, 4, and 5 DIP, BioGRID, and STRING are used as reference PPI network. We reported the best performance of N2V-MLP in the results. The proposed method, EssentialGIN, performs better than N2V-MLP in H. sapiens. Our intuition is that the high number of genes and more complicated interactions result in more complex architecture which is better represented by GIN (compared to N2V). A specific number of random walks in N2V, probably is not a sufficient tool to capture this structural information.

Since the implementation of EPGAT was not available, in the first part of experiments we used the results reported in [17], which was limited to AUROC, and compared AUROC in different spices using different PPI networks DIP, BioGRid and STRING. In our experiments, we used the same experiment setup and reported AUROC for EPGAT, DC, MLP, N2V-MLP (using only PPI, and also PPI+expression data), EPGAT (only PPI, PPI + expression, PPI + subcellular localization, PPI + orthology information) and also the proposed method, EssentialGIN, (only PPI, PPI + expression, PPI + subcellular localization, PPI + orthology information). The result is reported using DIP, BioGRid, and STRING in table 3-5 respectively. The best result in all the experiments regards to EssentialGIN + expression data.

In the following tables 3-8, "+expression" in the first column means (PPI + expression) and "+subcellular localization" means (PPI + expression+ subcellular localization).

**Table 3: Comparison of AUROC for essential gene prediction on three model organisms and H. sapiens using DIP PPI for degree centrality (DC), multilayer perceptron (MLP), Node2vec-MLP (N2V-MLP – fig 5), EPGAT [30] and proposed method (EssentialGIN).**

| | +E. coli | S. cerevisiae | D. melanogaster | H. sapiens |
|---|---|---|---|---|
| DC | 73.39(±0.05) | 68.07(±0.02) | 56.15(±0.09) | 59.86(±0.02) |
| MLP | 63.92(±0.06) | 58.95(±0.03) | 52.83(±0.09) | 53.86(±0.02) |
| N2V-NLP just using PPI | 65.02(±0.04) | 65.80(±0.01) | **52.92(±0.11)** | 64.08(±0.03) |
| + expression data | **85.62 (±0.03)** | 72.34 (±0.01) | 51.34(±0.08) | 82.63(±0.01) |
| EPGAT just PPI | 71.57(±0.83)) | 68.58(±1.41) | 38.61(±9.17) | 63.70(±2.05) |
| +expression data | 78.02(±1.14) | 72.32(±0.27) | 40.90(±3.68) | 79.63(±0.94) |
| +subcellular localization | _ _ _ _ _ | 72.57(±0.82) | 48.01(4.47) | 80.19(±0.93) |
| +orthology data | 79.96(±2.20) | 76.57(±0.74) | 32.90(±5.34) | 75.61(±0.90) |
| EssentialGIN just PPI | 65.99(±0.05) | 69.10 (±0.02) | 51.87(±0.07) | 64.97(±0.04)) |
| +expression data | 74.73 (±0.02) | 73.38 (±0.01) | **52.84(±0.08)** | **87.09(±0.01)** |
| +subcellular localization | _ _ __ | 73.41 (±0.01) | 52.07(±0.02) | 86.28(±0.01) |
| +orthology data | 78.21 (±0.01) | **73.32 (±0.01)** | 52.55(±0.02) | 81.02(±0.03) |

**Table 4: Comparison of AUROC for essential gene prediction on three model organisms and H. sapiens using BioGRid PPI for degree centrality (DC), multilayer perceptron (MLP), Node2vec-MLP (N2V-MLP – fig 5), EPGAT [30] and proposed method (EssentialGIN).**

| | E. coli | S. cerevisiae | D. melanogaster | H. sapiens |
|---|---|---|---|---|
| DC | 76.02(±0.04) | 77.33(±0.01) | 78.33(±0.04) | 82.06(±0.01) |
| MLP | 63.37(±0.06) | 65.13(±0.02) | 65.95(±0.05) | 77.01(±0.01) |
| N2v-MLP just PPI | 71.88(±0.05) | 76.84(±0.01) | 65.88(±0.05) | 81.03(±0.01) |
| +expression data | 78.15%(±0.02) | 76.75%(±0.01) | 60.36(±0.036) | 87.98(±0.03) |
| EPGAT just PPI | 89.90(±1.78) | 88.58(±1.41) | 77.20(±3.72) | 87.81(±0.62) |
| +expression data | 92.09(±0.97) | 89.02(±1.34) | 78.31(±2.22) | 88.04(±0.96) |
| +subcellular localization | _ _ _ _ _ | 88.71(±2.16) | **79.87(±1.87)** | 89.09(± 0.64) |
| +orthology data | **92.35(±1.15)** | 87.66(±2.58) | 78.78(±4.24) | 88.39(±0.52) |
| EssentialGIN just PPI | 63.06(±0.06) | 88.97(±0.01) | 67.92(±0.04) | 79.55(±0.02) |
| +expression data | 85.17 (±0.05) | 92.45 (±0.02) | 68.85 (±0.07) | 90.53(±0.00) |
| +subcellular localization | _ _ __ | 94.00 (±0.01) | 66.61 (±0.08) | **90.70(±0.00)** |
| +orthology data | 84.76(±0.02) | **94.34(±0.01)** | 60.47 (±0.06) | 88.80(±0.01) |

**Table 5: Comparison of AUROC for essential gene prediction on three model organisms and H. sapiens using STRING PPI for degree centrality (DC), multilayer perceptron (MLP), Node2vec-MLP (N2V-MLP – represented in fig 5), EPGAT [26] and proposed method (EssentialGIN).**

| | E. coli | S. cerevisiae | D. melanogaster | H. sapiens |
|---|---|---|---|---|
| DC | 82.92(±0.01) | 77.33(±0.01) | 68.17(±0.03) | 80.02(±0.01) |
| MLP | 73.09(±0.01) | 69.88(±0.02) | 65.19(±0.03) | 74.95(±0.01) |
| N2v-MLP just PPI | 91.29(±0.02) | 81.23(±0.01) | 72.13(±0.03) | 87.32(±0.00) |
| +expression data | 92.40(±0.20) | 81.*92*(±0.02) | 74.83 (±0.03) | 91.54(±0.01) |
| EPGAT just PPI | 96.43(±1.08) | **90.43(±0.45)** | 72.67(±2.89) | 90.34(±0.46) |
| +expression data | 96.71(±0.49) | 90.21(±0.57) | 73.71(±0.99) | 91.28(±0.42) |
| +subcellular localization | – – – – – | 90.01(±0.61) | 73.95(±3.07) | 91.52(±0.34) |
| +orthology data | **97.02(±0.50)** | 90.13(±1.08) | **77.82(±1.95)** | 90.95(±0.54) |
| EssentialGIN just PPI | 93.78(±0.01) | 87.73(±0.01) | 67.62(±0.06) | 89.39(±0.0) |
| +expression data | 95.61 (±0.02) | 88.64 (±0.02) | 68.57 (±0.04) | **92.89(±0.01)** |
| +subcellular localization | – – –– | 88.76(±0.01) | 66.77(±0.06) | 92.83(±0.01) |
| +orthology data | 95.02(±0.02) | 89.*46*(±0.02) | 63.93(±0.05) | 92.05(±0.0) |

In tables 3-6, we used AUROC measure in order to be able to compare our method to EPGAT also for which only AUROC reuslt were available. However as we will discuss later in this section, AUROC is not a perfect measure for this problem due to class-imbalance.

On E. coli data, EssentialGIN is not the best-performing method. For DIP network (table 3), N2V-MLP results in the highest AUROC value, and EPGAT is the best performing method for BioGRID (table 4) and STRING networks (table 5).

Best performance on S. cerevisiae data on DIP (table 3) and BioGRID (table 4) were achived using EssentialGIN. On STRING (table 5), the performance of EssentialGIN was close to best-performing method (EPGAT). For melanogaster however, EPGAT performed significantly better than other methods for BioGRID (table 4) and STRING (table) PPIs. For DIP (table 3), N2V-MLP was the best performing method, and the perfromance of EssntialGIN is close to best performing method.

In all the experiments using expression data in addition to PPI networ, has improved the result, however this is not the case for orthology data and subcellular localization data. Orthology data and subcellar localization data had very small impact on improveing the accuracy in most of the cases and even had a negative effect on the output quality in some cases (see table 5, D. melanogaster as an example).

In the next experiments (table 6-8), we reported AUROC, AUPR, precision, MCC, and accuracy on H. sapiens using different PPIs (DIP, BioGRid, and STRING respectively) for the previous methods, except EPGAT (since the results were not available).

**Table 6: AUROC, AUPR, precision, MCC and accuracy reported for H. sapiens essential gene prediction using DIP PPI network (+ other mentioned data); methods compared include degree centrality (DC), multilayer perceptron (MLP), Node2Vec-MLP (N2V-MLP, represented in fig 5),**

**and the proposed method, EssentialGIN. The effect of adding data such as gene expression data, subcellular localization and gene orthology data is also investigated.**

| H. sapiens | AUROC | AUPR | precision | MCC | accuracy |
|---|---|---|---|---|---|
| DC | 59.86(±0.02) | 29.31(±0.03) | 28.72(±0.03) | 9.58(±0.05) | 70.56(±0.02) |
| MLP | 53.86(±0.02) | 24.40(±0.01) | 24.82(±0.01) | 3.91(±0.04) | 61.77(±0.02) |
| N2v-MLP just PPI | 64.08(±0.03) | 34.15(±0.03) | 34.59(±0.03) | 17.42(±0.04) | 63.56(±0.02) |
| +expression data | 82.63(±0.01) | 55.30(±0.02) | 55.61(±0.02) | 42.66(±0.04) | 76.02(±0.02) |
| EssentialGIN just PPI | 64.97(±0.04) | 36.13(±0.03) | 36.55(±0.03) | 19.44(±0.06) | 68.23(±0.02) |
| +expression data | **87.09(±0.01)** | **63.53(±0.04)** | **63.86(±0.04)** | **50.07(±0.03)** | 78.53(±0.01) |
| +subcellular localization | 86.28(±0.01) | 62.10(±0.02) | 62.46(±0.02) | 49.72(±0.02) | 78.86(±0.01) |
| +orthology data | 81.02(±0.03) | 55.81(±0.05) | 56.14(±0.05) | 41.77(±0.04) | **79.50(±0.02)** |

**Table 7: AUROC, AUPR, precision, MCC and accuracy reported for H. sapiens essential gene prediction using BioGRid PPI network (+ other mentioned data); methods compared include degree centrality (DC), multilayer perceptron (MLP), Node2Vec-MLP (N2V-MLP, represented in fig 5), and the proposed method, EssentialGIN. The effect of adding data such as gene expression data, subcellular localization and gene orthology data is also investigated.**

| H. sapiens | AUROC | AUPR | precision | MCC | accuracy |
|---|---|---|---|---|---|
| DC | 82.06(±0.01) | 32.91(±0.02) | 33.04(±0.02) | 27.71(±0.02) | 86.98(±0.00) |
| MLP | 77.01(±0.01) | 28.76(±0.02) | 28.99(±0.02) | 28.82(±0.02) | 77.98(±0.01) |
| N2v-MLP just PPI | 81.03(±0.01) | 39.03(±0.02) | 39.26(±0.02) | 34.67(±0.02) | 78.91(±0.01) |
| +expression data | 87.98(±0.03) | 53.28(±0.06) | 53.41(±0.06) | 42.86(±0.07) | 78.73(±0.10) |
| EssentialGIN just PPI | 79.55(±0.02) | 32.66(±0.02) | 32.87(±0.02) | 31.99(±0.02) | 78.29(±0.01) |
| +expression data | 90.53(±0.00) | 57.38(±0.02) | 57.49(±0.02) | 47.89(±0.01) | 81.83(±0.01) |
| +subcellular localization | **90.70(±0.00)** | **57.81(±0.02)** | **57.92(±0.02)** | **48.40(±0.01)** | 82.69(±0.01) |
| +orthology data | 88.80(±0.01) | 56.39(±0.02) | 56.48(±0.02) | 46.99(±0.01) | **85.80(±0.01)** |

**Table 8: AUROC, AUPR, precision, MCC and accuracy reported for H. sapiens essential gene prediction using STRING PPI network (+ other mentioned data); methods compared include degree centrality (DC), multilayer perceptron (MLP), Node2Vec-MLP (N2V-MLP, represented in fig 5), and the proposed method, EssentialGIN. The effect of adding data such as gene expression data, subcellular localization and gene orthology data is also investigated.**

| H. sapiens | AUROC | AUPR | precision | MCC | accuracy |
|---|---|---|---|---|---|
| DC | 80.02(±0.01) | 31.00(±0.02) | 31.15(±0.02) | 29.19(±0.02) | 87.30(±0.00) |
| MLP | 74.95(±0.01) | 26.73(±0.02) | 26.94(±0.02) | 26.65(±0.02) | 77.68(±0.01) |
| N2v-MLP just PPI | 87.32(±0.00) | 54.38(±0.02) | 4.50(±0.02) | 44.00(±0.01) | 84.00(±0.01) |
| +expression data | 91.54(±0.01) | 62.96(±0.02) | 63.05(±0.02) | 49.95(±0.01) | 84.65(±0.01) |
| EssentialGIN just PPI | 89.39(±0.01) | 59.52(±0.02) | 59.60(±0.02) | 51.94(±0.01) | 88.34(±0.00) |
| +expression data | **92.89(±0.01)** | **68.61(±0.02)** | **68.66(±0.02)** | 55.72(±0.02) | 87.61(±0.00) |
| +subcellular localization | 92.83(±0.01) | 68.38(±0.02) | 68.43(±0.02) | 55.80(±0.01) | 87.86(±0.00) |
| +orthology data | 92.05(±0.00) | 66.28(±0.02) | 66.34(±0.01) | **56.65(±0.01)** | **89.52(±0.00)** |

As it can be observed in tables 6-8, for H. sapiens data the proposed method (EssentialGIN) demonstrates the best performance in all the measures (using PPI + gene expression data); The two best performing methods EssenialGIN and N2V-MLP had their best performance when STRING as the input PPI. We have to mention that the second best method (N2V-MLP) performs poorly in terms of AUPR, precison and recall compared to EssentialGIN.

The performance of EssentialGIN is also compared with SVM, gradient boosting, random forest, betweenness centrality and clustering coefficient for H. sapiens, and EssentialGIN outperforms all these methods in AUROC, AUPR, precision and recall (see appendix A, tables A1-A6). As it can be observed methods such as DC and betweenness centrality which are based on only structural features, perform poorly in terms of AUPR, precison and recall.

Finally we analysed some of the essential and non-essential nodes in order to recognize some local topological patterns. Nodes with large values of DC and BC contain both essential and non-essential nodes, however . We visualized the local structure of some essential/non-essential nodes in order to make an initial intuition about the local structure of these networks. We observed both categories of essential (figure 6) and non-essential (figure 7) genes with high or dense neihborhood. We could not find a correlation pattern between the density of 2- and 3-hop neighborhood nodes with gene essentiality in human data.  The same pattern were observed on other organisms.

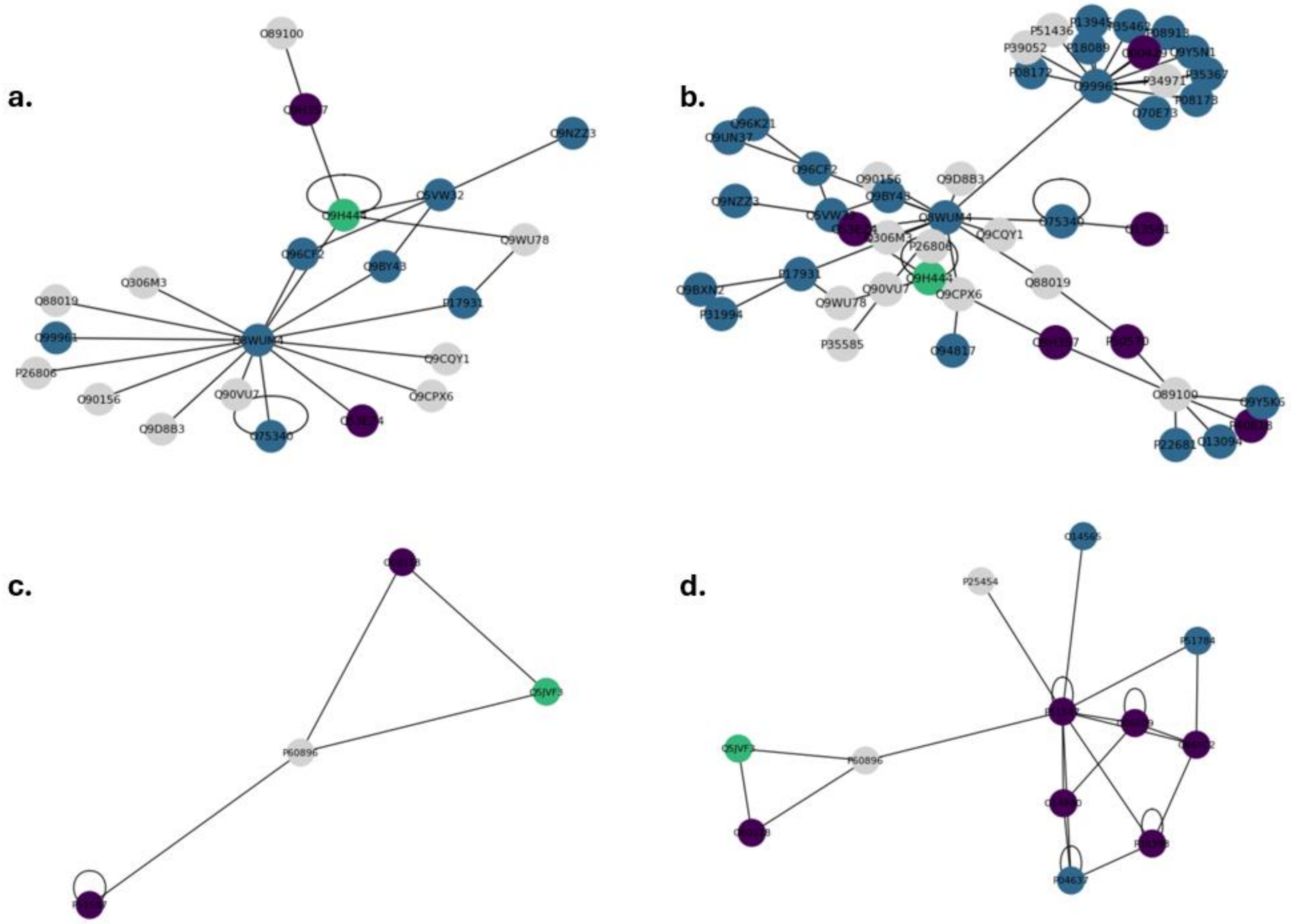

Figure 6. Local neighborhood of two essential proteins Q9H444 2-hop and 3-hop neighborhood (a and b) and Q5JVF3 2-hop and 3-hop neighborhood (c and d).

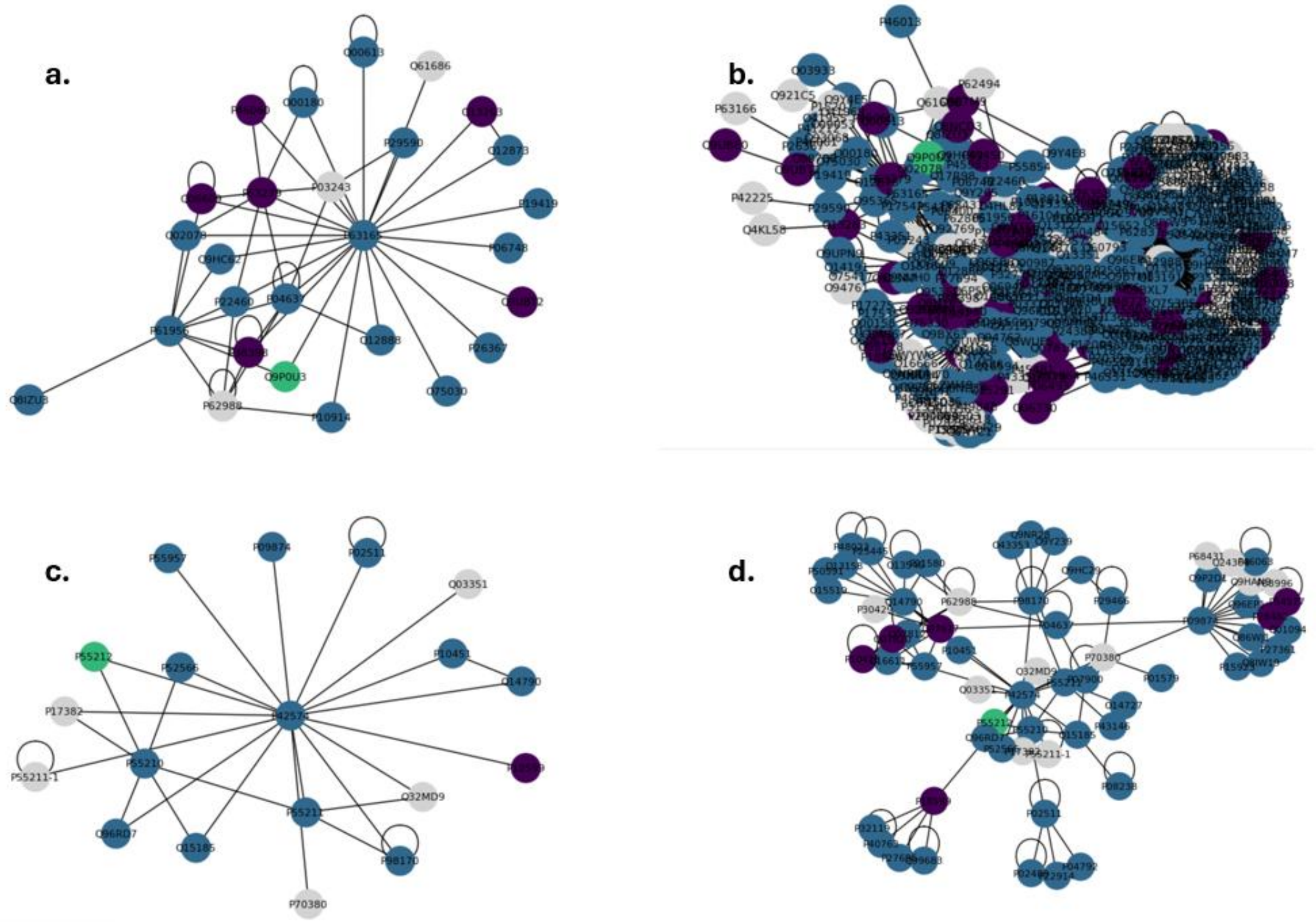

Figure 7. Local neighborhood of two non-essential proteins Q9P0U3 2-hop and 3-hop neighborhood (a and b) and P55212 2-hop and 3-hop neighborhood (c and d).

Finaly we used histogram analhysis. Histograms for degrees of non-essential and essential nodes (figure 8-9) shows similar distibution shapes for essential and non-essential nodes. However since the number of non-essential nodes is much more than the number of essential nodes, when considering a larger threshold for degree value, we noticed that the true positive rate (eq. 11) will be increase significantly compared to false positive rate (eq. 12).

$$TPR = \frac{TP}{TP + FN} \quad (11)$$

$$FPR = \frac{FP}{TN + FP} \quad (12)$$

Since the number of true negatives (non-essential nodes labeled as negative) is very large compared to false positives (non-essential nodes labeled as positive), FPR will be very low for larger degree values. As the result we can conclude that using AUROC is not a sutiable measure for this problem. We only used it since previous work also used it. But we can conlude that degree distiubtion is not distinguishing essential proteins from non-essential ones, and this can be also observed in the result available in the appendix in which degree centrality, betweenness centrality and clustering coefficient represent high values of AUROC, while the values of AUPR, precision and recall is significanlty lower using these algorithms for identifying essential genes.

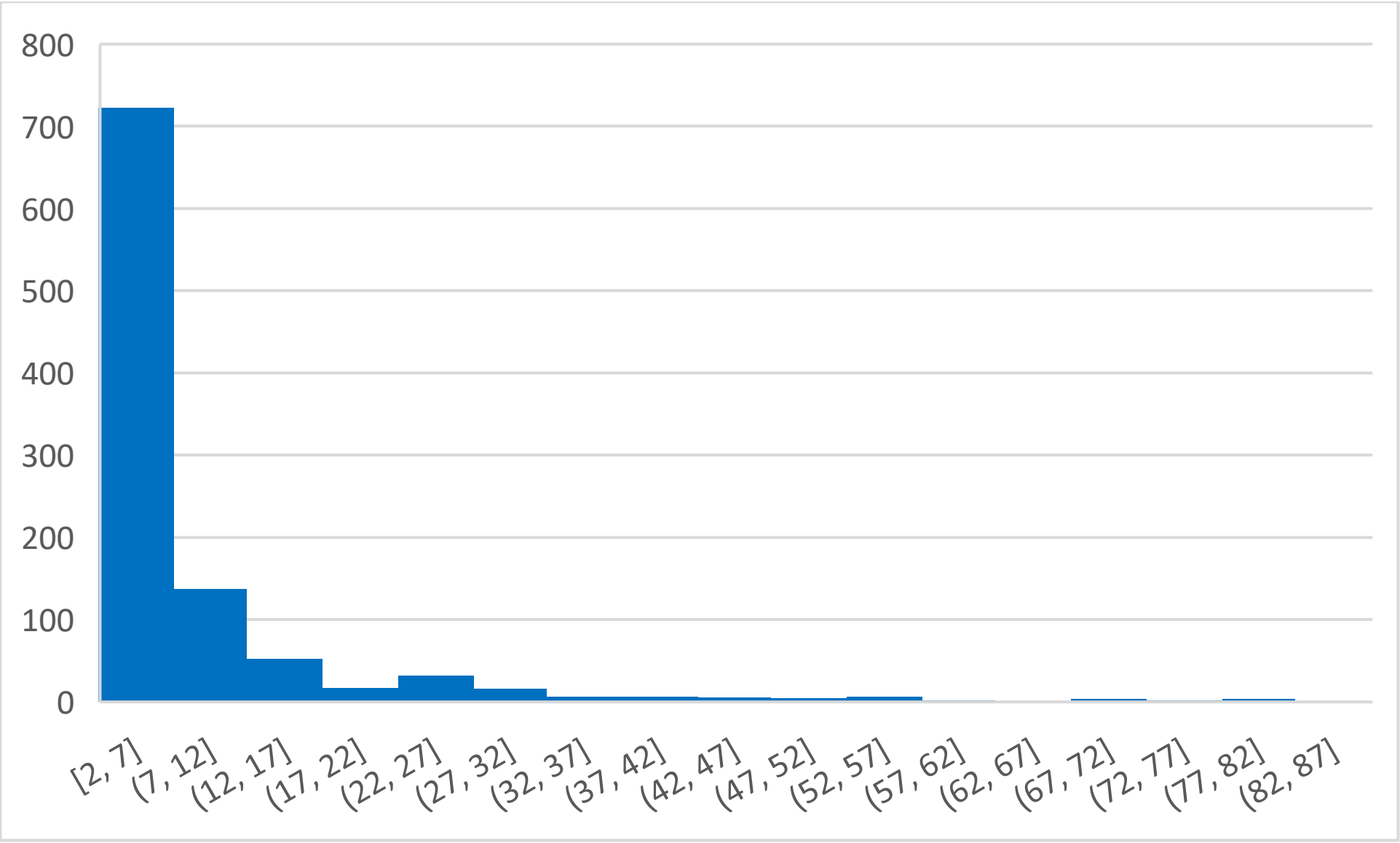

Figure 8. degree distribution in non-essential genes – reference netowrk: E. coli DIP. (four outliers are removed)

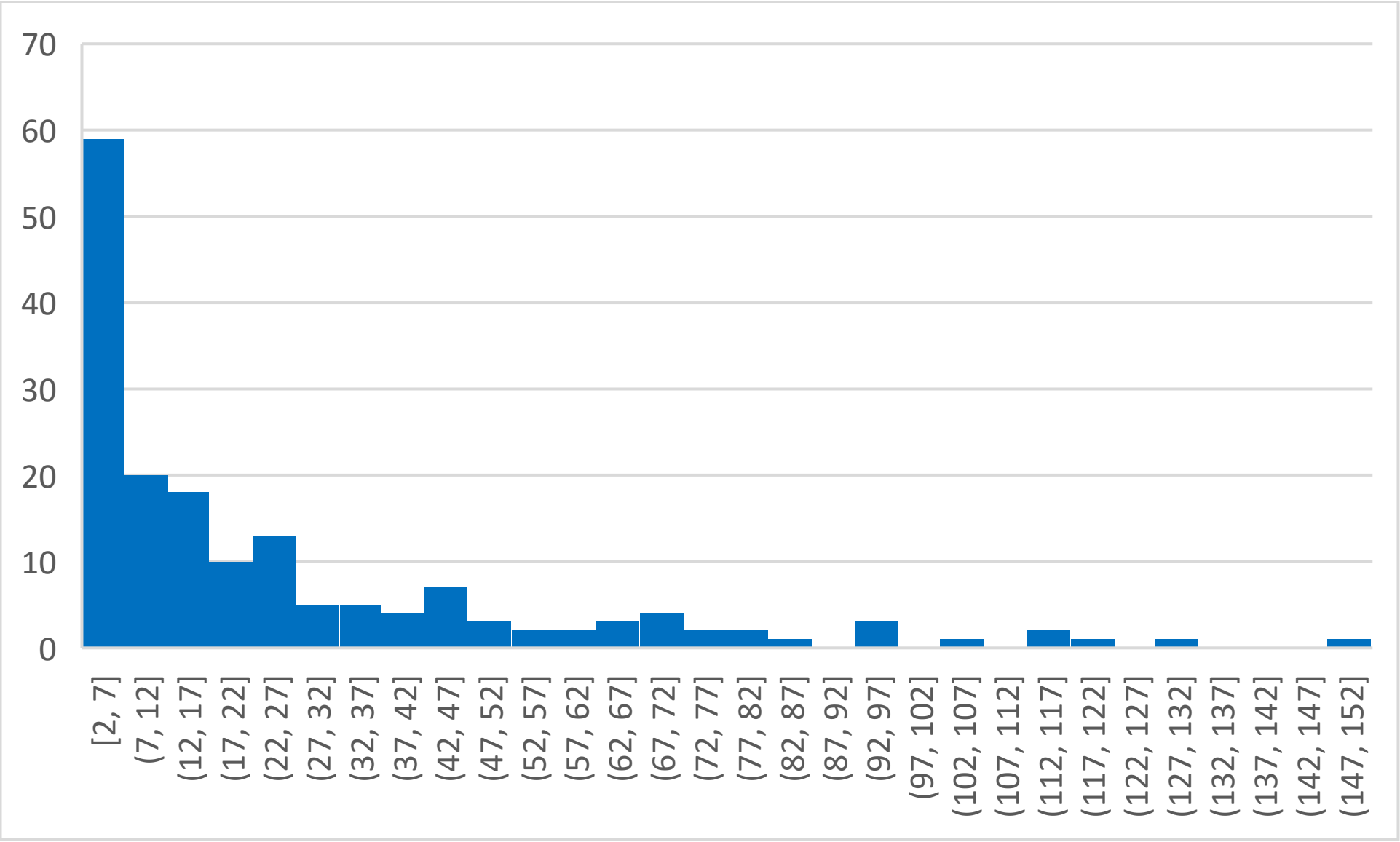

Figure 9. degree disribution in essential genes – reference netowrk: E. coli DIP.

We analyzed expression data for E. coli (the human expression file contains many missing values), and observed an obvious difference between the mean expression of samples in essential and non-essential genes; average 6.26 for all genes, average 27.98 for essential genes and 4.73 for non-essential genes. However, this factor cannot be seen as a discriminator between essential and non-essential genes.

As observed in our experiments, combining expressoin values with PPI increases the precision of different approaches. This mechanism can be further investigated and interpreted in order to better understand the strucutre and function of essential genes. Using GNNs in methods such as N2V-MLP, EssentialGIN and EPGAT is a good solution to integrate PPI and expression data, but extracting a pattern to explain gene essentiality using these features is not straight-forward using these methods as well.

## Conclusion

Centrality measures, which are measures to quantify structural properties of graph nodes, are the first computational method that are observed to be related with gene essentiality, however their predictive power is relatively low. In addition, integration of node attributes such as expression levels or subcellular localization with centrlity measures is not straightforward. Deep learning methods and specifically Graph Neural Networks (GNNs) are known as state-of-the-art models for predicting essential genes. The representation of nodes in GNNs is achieved through a message passing algorithms, using which the represenation of a node is updated using an integration of attributes of its neighbors (and itself). This representation is a function of node attributes, and considers node neighborhood information. While GNNs based on this message passing algorithm, are observed to be useful in gene essentiality prediction problem, they ignore the topological information of graph structure.

In this research, we used Graph Isomorphism networks (GINs), a type of GNN which is specifically defined to cover graph isomorphism. GINs are designed mainly to obtain graph embeddings; We modified original GIN model, in order to be able to use it for node classification. We designed a deep architecture based on GIN for predicting gene essentiality, hoping that both node attributes and topological structure of nodes together can improve the precision of gene essentiality prediction. GINs are mainly introduced for graph classification. With a slight modification, we used GINs for node classification. We integrated different gene attributes such as gene expression measurements, gene subcellular localization, and gene orthology information with the PPI network extracted features (GIN-based) in order to predict gene essentiality.

We have to mention here that other gene features for example features extracted from UniProt databse such as biological process, coding sequence information, domain, ligand, etc. are considered in existing research for discovering disease-gene associations [39]; these feartures also can be considered in predicting gene essentiality. In the current work, we followed the same approach and used cell-sublocalization data only so that we can compare our method with methods such as EPGAT. Our extensive experiments using different PPI networks (DIP, BioGRid, STRING), shows that the new appraoch (EssentialGIN) which is based on GINs, performs better than state-of-the-art methods including EPGAT (based on Graph Attention Networks), Node2Vec-MLP, centrality-based methods, random forest, gradient boosting  and MLP in H. sapiens on AUROC, AUPR, precision, and MCC. For E. coli, S. cerevisiae, and melanogaster our method is compared in terms of AUROC and is comparable with the best performing methods such as N2V-MLP and EPGAT.

## Acknowledgements

The authors thank the anonymous reviewers for their valuable suggestions.

## Author contributions

SMR developed and implemented the method. SMR, ZN conceptualized the idea and wrote the manuscript. PR verified the model and improved the architecture. NH validated the results, did the data extraction, protein id unifying and data verification. ZN supervised and administered the project. All authors read and approved the final manuscript.

## Funding

This research received no specific grant from any funding agency in the public, commercial, or not-for-profit sectors.

## Availability of data and materials

Source code is available at https://github.com/saharmansourirad/EssentialGIN.

## Declarations

**Ethics approval and consent to participate**

Not applicable.

**Consent for publication**

Not applicable.

**Competing interests**

The authors declare that they have no competing interests.

## References

[1] Y. T. Liang, H. Luo, Y. Lin, and F. Gao, "Recent advances in the characterization of essential genes and development of a database of essential genes," *Imeta,* vol. 3, no. 1, p. e157, 2024.

[2] X. Tang, J. Wang, J. Zhong, and Y. Pan, "Predicting essential proteins based on weighted degree centrality," *IEEE/ACM Transactions on Computational Biology and Bioinformatics,* vol. 11, no. 2, pp. 407-418, 2013.

[3] X. Zhang, W. Xiao, M. L. Acencio, N. Lemke, and X. Wang, "An ensemble framework for identifying essential proteins," *BMC bioinformatics,* vol. 17, no. 1, pp. 322-322, 2016, doi: 10.1186/s12859-016-1166-7.

[4] G. Li, M. Li, J. Wang, J. Wu, F.-X. Wu, and Y. Pan, "Predicting essential proteins based on subcellular localization, orthology and PPI networks," *BMC bioinformatics,* vol. 17, no. 8, pp. 279-279, 2016, doi: 10.1186/s12859-016-1115-5.
[5] G. Li, M. Li, J. Wang, Y. Li, and Y. Pan, "United Neighborhood Closeness Centrality and Orthology for Predicting Essential Proteins," *IEEE/ACM transactions on computational biology and bioinformatics,* vol. 17, no. 4, pp. 1451-1458, 2018, doi: 10.1109/tcbb.2018.2889978.
[6] X. Zhang, M. L. Acencio, and N. Lemke, "Predicting Essential Genes and Proteins Based on Machine Learning and Network Topological Features: A Comprehensive Review," *Frontiers in physiology,* vol. 7, no. NA, pp. 75-75, 2016, doi: 10.3389/fphys.2016.00617.
[7] F.-B. Guo *et al.*, "Accurate prediction of human essential genes using only nucleotide composition and association information," *Bioinformatics (Oxford, England),* vol. 33, no. 12, pp. 1758-1764, 2017, doi: 10.1093/bioinformatics/btx055.
[8] W. Liu, T. Tang, X. Lu, X. Fu, Y. Yang, and L. Peng, "MPCLCDA: predicting circRNA–disease associations by using automatically selected meta-path and contrastive learning," *Briefings in Bioinformatics,* vol. 24, no. 4, p. bbad227, 2023, doi: https://doi.org/10.1093/bib/bbad227
[9] S. Feng *et al.*, "Hypergraph models of biological networks to identify genes critical to pathogenic viral response," *BMC bioinformatics,* vol. 22, no. 1, p. 287, 2021, doi: https://doi.org/10.1186/s12859-021-04197-2
[10] S.-H. Wang *et al.*, "RFEM: A framework for essential microRNA identification in mice based on rotation forest and multiple feature fusion," *Computers in Biology and Medicine,* vol. 171, p. 108177, 2024, doi: https://doi.org/10.1016/j.compbiomed.2024.108177
[11] H. Öztürk, A. Özgür, and E. Ozkirimli, "DeepDTA: deep drug–target binding affinity prediction," *Bioinformatics (Oxford, England),* vol. 34, no. 17, pp. i821-i829, 2018, doi: 10.1093/bioinformatics/bty593.
[12] M. Zeng, M. Li, F.-X. Wu, Y. Li, and Y. Pan, "DeepEP: a deep learning framework for identifying essential proteins," *BMC bioinformatics,* vol. 20, no. 16, pp. 1-10, 2019, doi: 10.1186/s12859-019-3076-y.
[13] M. Zeng *et al.*, "A Deep Learning Framework for Identifying Essential Proteins by Integrating Multiple Types of Biological Information," *IEEE/ACM transactions on computational biology and bioinformatics,* vol. 18, no. 1, pp. 296-305, 2021, doi: 10.1109/tcbb.2019.2897679.
[14] L. S. Hasan MA, "DEEPLYESSENTIAL: A deep neural network for predicting essential genes in microbes.," *BioRxiv,* 2019.
[15] Weng W, INet XZ. Convolutional networks for biomedical image segmentation., 2021, 9. DOI: https://doi. org/10.1109/ACCESS. 2021:16591-603.
[16] A. Grover and J. Leskovec, "KDD - node2vec: Scalable Feature Learning for Networks," *KDD : proceedings. International Conference on Knowledge Discovery & Data Mining,* vol. 2016, no. NA, pp. 855-864, 2016, doi: 10.1145/2939672.2939754.
[17] Y.-Y. Zhang, D.-M. Liang, and P.-F. Du, "iEssLnc: quantitative estimation of lncRNA gene essentialities with meta-path-guided random walks on the lncRNA-protein interaction network," *Briefings in Bioinformatics,* vol. 24, no. 3, p. 1-12, 2023, doi: https://doi.org/10.1093/bib/bbad097
[18] N. Wang, M. Zeng, J. Zhang, Y. Li, and M. Li, "Ess-NEXG: Predict Essential Proteins by Constructing a Weighted Protein Interaction Network Based on Node Embedding and XGBoost," in *Bioinformatics Research and Applications*, Cham, Z. Cai, I. Mandoiu, G. Narasimhan, P. Skums, and X. Guo, Eds., 2020// 2020: Springer International Publishing, pp. 95-104.
[19] X. Zhang, W. Xiao, and W. Xiao, "DeepHE: Accurately Predicting Human Essential Genes based on Deep Learning," *NA,* vol. NA, no. NA, pp. NA-NA, 2020, doi: 10.1101/2020.02.14.950048.
[20] J. Schapke, A. Tavares, and M. Recamonde-Mendoza, "EPGAT: Gene Essentiality Prediction With Graph Attention Networks," *IEEE/ACM Trans Comput Biol Bioinform,* vol. PP, Jan 26 2021, doi: 10.1109/TCBB.2021.3054738.
[21] P. Veličković, G. Cucurull, A. Casanova, A. Romero, P. Lio, Y. Bengio. Graph attention networks. arXiv preprint arXiv:1710.10903. 2017 Oct 30..

[22] T. Kipf and M. Welling, " Semi-supervised classification with graph convolutional networks" . arXiv preprint arXiv:1609.02907. 2016 Sep 9.

[23] A. Vaswani *et al.*, "NIPS - Attention is All you Need," Advances in Neural Information Processing Systems. , pp. 5998-6008, 2017, doi: NA.

[24] X. Peng, J. Wang, J. Wang, F.-X. Wu, and Y. Pan, "Rechecking the Centrality-Lethality Rule in the Scope of Protein Subcellular Localization Interaction Networks," *PloS one,* vol. 10, no. 6, pp. e0130743-NA, 2015, doi: 10.1371/journal.pone.0130743.

[25] R. Hasibi, "Deep Learning and Deep Reinforcement Learning for Graph Based Applications," 2024.

[26] K. X. W. H. J. Leskovec and S. Jegelka, "How powerful are graph neural networks," *ICLR. Keyulu Xu Weihua Hu Jure Leskovec and Stefanie Jegelka,* 2019.

[27] K. Xu, W. Hu, J. Leskovec, and S. Jegelka, "How Powerful are Graph Neural Networks?," in *International Conference on Learning Representations*, 2018.

[28] W. Hu *et al.*, "Strategies for pre-training graph neural networks," *arXiv preprint arXiv:1905.12265,* 2019.

[29] W.-H. Chen, G. Lu, X. Chen, X.-M. Zhao, and P. Bork, "OGEE v2: an update of the online gene essentiality database with special focus on differentially essential genes in human cancer cell lines," *Nucleic acids research,* vol. 45, no. D1, pp. D940-D944, 2016, doi: 10.1093/nar/gkw1013.

[30] L. Salwinski, C. S. Miller, A. J. Smith, F. K. Pettit, J. U. Bowie, and D. Eisenberg, "The Database of Interacting Proteins: 2004 update," *Nucleic Acids Research,* vol. 32, no. 1, pp. 239-241, 2001, doi: doi: 10.1093/nar/gkh086

[31] A. E. Carpenter *et al.*, "CellProfiler: image analysis software for identifying and quantifying cell phenotypes," *Genome biology,* vol. 7, pp. 1-11, 2006, doi:10.1186/gb-2006-7-10-r100

[32] D. Szklarczyk *et al.*, "STRING v11: protein–protein association networks with increased coverage, supporting functional discovery in genome-wide experimental datasets," *Nucleic acids research,* vol. 47, no. D1, pp. D607-D613, 2019, doi: 10.1093/nar/gky1131

[33] C. Peng and F. Gao, "Protein localization analysis of essential genes in prokaryotes," *Scientific reports,* vol. 4, no. 1, pp. 1-7, 2014, doi: 10.1038/srep06001

[34] J. X. Binder *et al.*, "COMPARTMENTS: unification and visualization of protein subcellular localization evidence," *Database : the journal of biological databases and curation,* vol. 2014, no. 2014, pp. bau012-bau012, 2014, doi: 10.1093/database/bau012.

[35] M. Gennarelli and A. Cattaneo, "Genetic Variations and Association," *International Review of Neurobiology,* vol. 94, pp. 129-151, 2010.

[36] E. L. L. Sonnhammer and G. Östlund, "InParanoid 8: orthology analysis between 273 proteomes, mostly eukaryotic," *Nucleic acids research,* vol. 43, no. Database issue, pp. 234-239, 2014, doi: 10.1093/nar/gku1203.

[37] K. A. Carlson *et al.*, "Genome-wide gene expression in relation to age in large laboratory cohorts of Drosophila melanogaster," *Genetics research international,* vol. 2015, 2015, doi: http://dx.doi.org/10.1155/2015/835624

[38] T. Barrett *et al.*, "NCBI GEO: archive for functional genomics data sets—update," *Nucleic acids research,* vol. 41, no. D1, pp. D991-D995, 2012, doi: https://doi.org/10.1093/nar/gks1193

[39] S. K. Ata, L. Ou-Yang, Y. Fang, C.-K. Kwoh, M. Wu, and X.-L. Li, "Integrating node embeddings and biological annotations for genes to predict disease-gene associations," *BMC systems biology,* vol. 12, pp. 31-44, 2018, doi: https://doi.org/10.1186/s12918-018-0662-y

Appendix A

**Table A1: AUROC, AUPR, precision, MCC and accuracy reported for H. sapiens essential gene prediction using DIP PPI network (+ other mentioned data); methods compared include betweenness centrality, clustering coefficients, Random Forest, Support Vector Machines (SVM), Gradient Boosting and the proposed method, EssentialGIN. The effect of adding data such as gene expression data, subcellular localization and gene orthology data is also investigated.**

| H. sapiens | AUROC | AUPR | precision | MCC | accuracy |
|---|---|---|---|---|---|
| betweenness centrality | 57.34(±0.02) | 27.62(±0.02) | 26.93(±0.02) | 8.35(±0.04) | 70.28(±0.01) |
| clustering coefficients | 60.91(±0.01) | 30.83(±0.03) | 27.77(±0.02) | 16.40(±0.04) | 73.18(±0.02) |
| Random forest just using PPI | 52.16 (±0.02) | 22.78 (±0.02) | 23.17 (±0.02) | 1.16 (±0.04) | 78.46 (±0.00) |
| + expression data | 87.91 (±0.01) | 56.38 (±0.02) | 56.39 (±0.02) | 36.53 (±0.03) | 81.80 (±0.01) |
| +subcellular localization | 82.79 (±0.01) | 57.57 (±0.02) | 57.40 (±0.02) | 36.26 (±0.04) | 81.80 (±0.01) |
| +orthology data | 80.14 (±0.02) | 57.76 (±0.03) | 57.47 (±0.03) | 38.82 (±0.03) | 82.18 (±0.01) |
| SVM just PPI | 53.25 (±0.03) | 24.09 (±0.02) | 24.51(±0.02) | 0.00 (±0.00) | 78.49 (±0.00) |
| +expression data | 84.38 (±0.01) | 56.21(±0.03) | 56.58 (±0.03) | 38.33 (±0.04) | 81.54 (±0.01) |
| +subcellular localization | 83.54 (±0.02) | 56.52 (±0.04) | 56.82 (±0.04) | 36.38 (±0.05) | 81.48 (±0.01) |
| +orthology data | 80.00 (±0.02) | 56.65 (±0.02) | 56.90 (±0.02) | 37.08 (±0.03) | 81.87 (±0.01) |
| Gradient Boosting just PPI | 54.30(±0.03) | 23.83 (±0.02) | 24.21(±0.02) | -1.34(±0.03) | 77.39(±0.01) |
| +expression data | 81.91(±0.01) | 54.16(±0.03) | 54.45(±0.03) | 33.48(±0.03) | 81.08(±0.01) |
| +subcellular localization | 82.54(±0.01) | 55.37(±0.02) | 55.70(±0.02) | 37.04(±0.03) | 81.81(±0.01) |
| +orthology data | 83.34(±0.01) | 59.83(±0.03) | 60.05(±0.03) | 40.86(±0.02) | **82.51(±0.01)** |
| ESSENTIALGIN just PPI | 64.97(±0.04) | 36.13(±0.03) | 36.55(±0.03) | 19.44(±0.06) | 68.23(±0.02) |
| +expression data | **87.09(±0.01)** | **63.53(±0.04)** | **63.86(±0.04)** | **50.07(±0.03)** | 78.53(±0.01) |
| +subcellular localization | 86.28(±0.01) | 62.10(±0.02) | 62.46(±0.02) | 49.72(±0.02) | 78.86(±0.01) |
| +orthology data | 81.02(±0.03) | 55.81(±0.05) | 56.14(±0.05) | 41.77(±0.04) | 79.50(±0.02) |

**Table A2: AUROC, AUPR, precision, MCC and accuracy reported for H. sapiens essential gene prediction using BioGRid PPI network (+ other mentioned data); methods compared include betweenness centrality, clustering coefficients, Random Forest, Support Vector Machines (SVM), Gradient Boosting and the proposed method, EssentialGIN. The effect of adding data such as gene expression data, subcellular localization and gene orthology data is also investigated.**

| H. sapiens | AUROC | AUPR | precision | MCC | accuracy |
|---|---|---|---|---|---|
| betweenness centrality | 75.79(±0.01) | 24.86(±0.01) | 25.00(±0.01) | 18.77(±0.02) | 85.40(±0.00) |
| clustering coefficients | 68.54(±0.01) | 15.72(±0.00) | 15.74(±0.00) | 2.93(±0.01) | 81.89(±0.00) |
| Random forest just using PPI | 78.88 (±0.01) | 29.80 (±0.01) | 29.65 (±0.01) | 5.51 (±0.02) | 89.03 (±0.00) |
| + expression data | 88.14 (±0.01) | 52.14 (±0.02) | 51.76 (±0.02) | 35.42 (±0.02) | 90.32 (±0.00) |
| +subcellular localization | 88.44 (±0.01) | 53.79 (±0.02) | 53.36 (±0.02) | 35.49 (±0.02) | 90.37 (±0.02) |
| +orthology data | 87.40 (±0.01) | 55.34(±0.01) | 54.89 (±0.01) | 40.62 (±0.01) | 90.83 (±0.00) |
| SVM just PPI | 70.87 (±0.01) | 26.05 (±0.02) | 26.31 (±0.02) | 10.35 (±0.04) | 88.90 (±0.00) |
| +expression data | 86.14 (±0.01) | 49.42 (±0.01) | 49.54 (±0.01) | 30.51 (±0.03) | 89.79 (±0.00) |
| +subcellular localization | 86.55 (±0.01) | 49.99 (±0.02) | 50.10 (±0.02) | 32.05 (±0.02) | 89.88 (±0.00) |
| +orthology data | 84.78 (±0.01) | 52.58 (±0.01) | 52.69 (±0.01) | 40.83 (±0.01) | 90.79 (±0.00) |
| Gradient Boosting just PPI | 81.28(±0.01) | 31.42(±0.02) | 31.66(±0.02) | 10.27(±0.03) | 88.79(±0.00) |
| +expression data | 88.52(±0.01) | 52.12(±0.03) | 52.29(±0.03) | 35.29(±0.02) | 90.21(±0.00) |
| +subcellular localization | 88.86(±0.00) | 53.32(±0.02) | 53.45(±0.02) | 37.39(±0.02) | 90.36(±0.00) |
| +orthology data | 89.18(±0.00) | 56.18(±0.02) | 56.31(±0.02) | 44.79(±0.02) | **91.04(±0.00)** |
| ESSENTIALGIN just PPI | 79.55(±0.02) | 32.66(±0.02) | 32.87(±0.02) | 31.99(±0.02) | 78.29(±0.01) |
| +expression data | 90.53(±0.00) | 57.38(±0.02) | 57.49(±0.02) | 47.89(±0.01) | 81.83(±0.01) |
| +subcellular localization | **90.70(±0.00)** | **57.81(±0.02)** | **57.92(±0.02)** | **48.40(±0.01)** | 82.69(±0.01) |
| +orthology data | 88.80(±0.01) | 56.39(±0.02) | 56.48(±0.02) | 46.99(±0.01) | 85.80(±0.01) |

**Table A3: AUROC, AUPR, precision, MCC and accuracy reported for H. sapiens essential gene prediction using STRING PPI network (+ other mentioned data); methods compared include betweenness centrality, clustering coefficients, Random Forest, Support Vector Machines (SVM), Gradient Boosting and the proposed method, EssentialGIN. The effect of adding data such as gene expression data, subcellular localization and gene orthology data is also investigated.**

| H. sapiens | AUROC | AUPR | precision | MCC | accuracy |
|---|---|---|---|---|---|
| betweenness centrality | 73.39(±0.05) | 68.07(±0.02) | 56.15(±0.09) | 59.86(±0.02) | 85.88% |
| clustering coefficients | 63.25(±0.01) | 13.19(±0.00) | 13.25(±0.00) | -0.50(±0.01) | 82.25(±0.00) |
| Random forest just using PPI | 76.11 (±0.01) | 27.53 (±0.02) | 27.27 (±0.02) | 4.16 (±0.03) | 89.92 (±0.00) |
| + expression data | 88.60 (±0.01) | 55.70 (±0.03) | 55.33 (±0.03) | 42.30 (±0.03) | 91.70 (±0.00) |
| +subcellular localization | 89.14 (±0.01) | 58.44 (±0.03) | 34.59(±0.03) | 17.42(±0.04) | 63.56(±0.02) |
| +orthology data | 82.63(±0.01) | 55.30(±0.02) | 57.94 (±0.03) | 42.15 (±0.03) | 91.74 (±0.00) |
| SVM just PPI | 70.52 (±0.01) | 24.44 (±0.02) | 24.64 (±0.02) | 7.38 (±0.02) | 89.84 (±0.00) |
| +expression data | 86.49 (±0.01) | 54.07 (±0.03) | 54.16 (±0.03) | 41.02 (±0.03) | 91.41 (±0.00) |
| +subcellular localization | 86.70 (±0.01) | 55.31 (±0.03) | 55.39 (±0.03) | 43.51 (±0.02) | 91.68 (±0.00) |
| +orthology data | 86.14 (±0.01) | 55.16 (±0.03) | 55.26 (±0.03) | 43.85 (±0.03) | 91.81 (±0.00) |
| Gradient Boosting just PPI | 79.06(±0.01) | 28.69(±0.02) | 28.89(±0.02) | 7.78(±0.03) | 89.71(±0.00) |
| +expression data | 89.52(±0.01) | 55.62(±0.03) | 55.83(±0.03) | 42.89(±0.03) | 91.68(±0.00) |
| +subcellular localization | 89.88(±0.01) | 56.45(±0.03) | 56.55(±0.03) | 44.00(±0.03) | 91.67(±0.00) |
| +orthology data | **90.14(±0.01)** | 58.58(±0.03) | 58.74(±0.03) | 46.74(±0.03) | **91.98(±0.00)** |
| ESSENTIALGIN just PPI | 64.97(±0.04) | 36.13(±0.03) | 36.55(±0.03) | 19.44(±0.06) | 68.23(±0.02) |
| +expression data | 87.09(±0.01) | **63.53(±0.04)** | **63.86(±0.04)** | **50.07(±0.03)** | 78.53(±0.01) |
| +subcellular localization | 86.28(±0.01) | 62.10(±0.02) | 62.46(±0.02) | 49.72(±0.02) | 78.86(±0.01) |
| +orthology data | 81.02(±0.03) | 55.81(±0.05) | 56.14(±0.05) | 41.77(±0.04) | 79.50(±0.02) |

**Table A4: Comparison of AUROC for essential gene prediction on three model organisms and H. sapiens using DIP PPI for betweenness centrality, clustering coefficients, Random Forest, Support Vector Machines (SVM) and Gradient Boosting.**

| | E. coli | S. cerevisiae | D. melanogaster | H. sapiens |
|---|---|---|---|---|
| betweenness centrality | 72.75(±0.05) | 65.61(±0.00) | 53.98(±0.09) | 57.34(±0.02) |
| clustering coefficients | 66.18(±0.02) | 67.01(±0.02) | 46.54(±0.04) | 60.91(±0.01) |
| Random forest just using PPI | 69.08(±0.05) | 63.77(±0.02) | 51.40(±0.07) | 52.16(±0.02) |
| + expression data | 87.91(±0.02) | 72.29 (±0.02) | **58.00 (±0.08)** | 82.09(±0.01) |
| +subcellular localization | – – – – – | 72.48 (±0.01) | 57.95 (±0.09) | 82.79(±0.01) |
| +orthology data | 91.07(±0.02) | 79.38 (±0.01) | 42.99 (±0.09) | 80.14(±0.02) |
| SVM just PPI | 67.14(±0.04) | 59.19(±0.03) | 49.50(±0.06) | 53.25(±0.03) |
| +expression data | 82.58 (±0.03) | 68.10 (±0.02) | 52.54 (±0.12) | **84.38 (±0.01)** |
| +subcellular localization | – – –– | 68.11 (±0.02) | 48.02 (±0.10) | 83.54(±0.02) |
| +orthology data | 86.49 (±0.04) | 78.24 (±0.01) | 55.34 (±0.09) | 80.00(±0.02) |
| Gradient Boosting just PPI | 70.20(±0.04) | 64.85(±0.02) | 49.63(±0.09) | 54.30 (±0.03) |
| +expression data | 88.42 (±0.02) | 73.38 (±0.01) | 52.84 (±0.08) | 81.91(±0.01) |
| +subcellular localization | – – –– | 73.41 (±0.01) | 52.07 (±0.07) | 82.54(±0.01) |
| +orthology data | **91.53 (±0.01)** | **81.43 (±0.01)** | 42.68 (±0.07) | 83.34(±0.01) |

**Table A5: Comparison of AUROC for essential gene prediction on three model organisms and H. sapiens using BioGrid PPI for betweenness centrality, clustering coefficients, Random Forest, Support Vector Machines (SVM) and Gradient Boosting.**

| | E. coli | S. cerevisiae | D. melanogaster | H. sapiens |
|---|---|---|---|---|
| betweenness centrality | 73.94(±0.04) | 60.06(±0.02) | 68.27(±0.04) | 75.79(±0.01) |
| clustering coefficients | 42.55(±0.03) | 55.42(±0.02) | 71.45(±0.02) | 68.54(±0.01) |
| Random forest just using PPI | 71.31(±0.03) | 68.97(±0.02) | 64.60(±0.08) | 78.88(±0.01) |
| + expression data | 77.18 (±0.03) | 75.23 (±0.02) | 64.71 (±0.05) | 88.14(±0.01) |
| +subcellular localization | – – – – – | 75.49 (±0.02) | 67.01 (±0.06) | 88.44(±0.01) |
| +orthology data | 80.78 (±0.03) | 81.67 (±0.02) | 70.04 (±0.08) | 87.40(±0.01) |
| SVM just PPI | 63.40(±0.04) | 64.20(±0.01) | 69.76(±0.07) | 70.87(±0.01) |
| +expression data | 69.88 (±0.05) | 70.32 (±0.02) | 61.11 (±0.06) | 86.14(±0.01) |
| +subcellular localization | – – –– | 70.32 (±0.02) | 59.01 (±0.06) | 86.55(±0.01) |
| +orthology data | 71.75 (±0.04) | 78.93 (±0.02) | 64.64 (±0.04) | 84.78(±0.01) |
| Gradient Boosting just PPI | 74.62(±0.04) | 70.08(±0.02) | 72.39(±0.07) | 81.28(±0.01) |
| +expression data | 81.44 (±0.03) | 76.17 (±0.02) | **75.51 (±0.05)** | 88.52(±0.01) |
| +subcellular localization | – – –– | 76.20 (±0.02) | 75.37 (±0.04) | 88.86(±0.00) |
| +orthology data | **84.76 (±0.02)** | **83.00 (±0.01)** | 73.95 (±0.04) | **89.18(±0.00)** |

**Table A6: Comparison of AUROC for essential gene prediction on three model organisms and H. sapiens using STRING PPI for betweenness centrality, clustering coefficients, Random Forest, Support Vector Machines (SVM) and Gradient Boosting.**

| | E. coli | S. cerevisiae | D. melanogaster | H. sapiens |
|---|---|---|---|---|
| betweenness centrality | 67.64(±0.02) | 66.92(±0.01) | 64.88(±0.04) | 59.86(±0.02) |
| clustering coefficients | 55.58(±0.04) | 63.63(±0.02) | 51.03(±0.04) | 63.25(±0.01) |
| Random forest just using PPI | 74.45 (±0.02) | 73.58 (±0.02) | 65.75 (±0.06) | 76.11 (±0.01) |
| + expression data | 80.74 (±0.02) | 75.65 (±0.01) | 59.25 (±0.03) | 88.60 (±0.01) |
| +subcellular localization | – – – – – | 75.95 (±0.01) | 57.98 (±0.05) | 89.14 (±0.01) |
| +orthology data | 80.82 (±0.03) | 81.17 (±0.01) | 63.30 (±0.04) | 88.60 (±0.01) |
| SVM just PPI | 74.60 (±0.03) | 69.26 (±0.02) | 60.70 (±0.04) | 70.52 (±0.01) |
| +expression data | 77.68 (±0.03) | 73.69 (±0.01) | 58.82 (±0.04) | 86.49 (±0.01) |
| +subcellular localization | – – – – – | 73.69 (±0.00) | 57.62 (±0.04) | 86.70 (±0.01) |
| +orthology data | 76.09 (±0.04) | 78.88 (±0.01) | 58.03 (±0.05) | 86.14 (±0.01) |
| Gradient Boosting just PPI | 79.59(±0.02) | 75.49(±0.02) | 66.77(±0.04) | 79.06(±0.01) |
| +expression data | 83.87 (±0.02) | 77.88 (±0.01) | 68.47 (±0.04) | 89.52(±0.01) |
| +subcellular localization | – – –– | 77.99 (±0.01) | 69.50 (±0.03) | 89.88(±0.01) |
| +orthology data | **85.17 (±0.01)** | **82.79 (±0.01)** | **69.83 (±0.05)** | **90.14(±0.01)** |